# Explainability of Point Cloud Neural Networks Using SMILE: Statistical Model-Agnostic Interpretability with Local Explanations

**Seyed Mohammad Ahmadi [a*], Koorosh Aslansefat [b], Rubén Valcarce-Diñeiro [a], Joshua Barnfather [c]**

[a] Centre of Excellence for Data Science, Artificial Intelligence and Modelling (DAIM), Faculty of Science and Engineering, University of Hull, Hull, U.K.

[b] School of Computer Science, Faculty of Science and Engineering, University of Hull, Hull, U.K.

[c] Dock Robotics Ltd, Hull, U.K.

## Abstract

In today's world, the significance of explainable AI (XAI) is growing in robotics and point cloud applications, as the lack of transparency in decision-making can pose considerable safety risks, particularly in autonomous systems. As these technologies are integrated into real-world environments, ensuring that model decisions are interpretable and trustworthy is vital for operational reliability and safety assurance. This study explores the implementation of SMILE, a novel explainability method originally designed for deep neural networks, on point cloud-based models. SMILE builds on LIME by incorporating Empirical Cumulative Distribution Function (ECDF) statistical distances, offering enhanced robustness and interpretability, particularly when the Anderson-Darling distance is used. The approach demonstrates superior performance in terms of fidelity loss, $R^2$ scores, and robustness across various kernel widths, perturbation numbers, and clustering configurations. Moreover, this study introduces a stability analysis for point cloud data using the Jaccard index, establishing a new benchmark and baseline for model stability in this field. The study further identifies dataset biases in the classification of the 'person' category, emphasizing the necessity for more comprehensive datasets in safety-critical applications like autonomous driving and robotics. The results underscore the potential of advanced explainability models and highlight areas for future research, including the application of alternative surrogate models and explainability techniques in point cloud data.

\* Corresponding author at: University of Hull, U.K.,
Email addresses: seyed-mohammad.ahmadi-2022@hull.ac.uk (S.M. Ahmadi), k.aslansefat@hull.ac.uk (K. Aslansefat), r.valcarce-dineiro@hull.ac.uk (R. Valcarce-Diñeiroa), joshua.barnfather@dock-robotics.com (J. Barnfather)

# 1. Introduction

Since the invention of the first machine, humankind has long aspired to create a fully automated world where all tasks are performed by machines. Recently, advancements in artificial intelligence, particularly with large language models like ChatGPT, which can communicate and respond in human language, have brought humankind closer than ever to realizing this long-held dream. However, achieving this vision requires another critical advancement: the development of machines capable of performing physical activities. Progress in the field of robotics appears to be the gateway to entering a fully automated world as the next step.

International Federation of Robotics (IFR) reported a record high of 517,385 new industrial robots installed in factories worldwide in 2021, with over 3.5 million units currently performing industrial tasks [1]. The advancement of robotics requires improved visual interpretation and 3D object representation techniques to enable robots to accurately perceive and interact with their environments.

Voxel grids, 3D meshes, multi-view camera projections and point clouds are the four primary forms of 3D objects representations [2]. A point cloud is a set of data points defined in a 3D coordinate system, which has recently gained more attention from the scientific community and is employed in a variety of areas, including 3D shape classification, 3D segmentations and object detection and tracking. This is because point clouds are not only easier to collect using LiDAR and RGB-D sensors but also provide a more thorough representation of objects compared to other forms. Moreover, point cloud data can efficiently store various attributes such as temperature, colour, and other specific parameters of an object at each point [3]. Most of the point cloud segmentation and classification techniques rely on deep neural networks (DNNs). However, these methods are more complicated and time-consuming than typical 2D DNNs used for images. This complexity arises from the disordered nature of point clouds, meaning there is no inherent order between data structures and spatial coordinates in the neighbourhood [2].

Until now, various methods are used for point cloud classification, categorised based on their feature learning architectures. The most notable ones include continuous and discrete convolution-based methods, Graph-based methods, pointwise MLP methods, and hierarchical data structure-based methods [2-4]. These methods have used different datasets with varying numbers and types of classes, some synthetic and others from the real world. Their representations include point clouds and other types like RGB-D and mesh, which are converted into point clouds before use [5-11].

Nevertheless, most DNN methods are not transparent and are often regarded as a black box, despite their satisfactory performance in classification tasks. On the other hand, the use of point cloud data in expensive and critical applications such as robotics and autonomous driving has become increasingly common. Therefore, the low decision confidence and poor interpretability of these models can significantly jeopardize people's lives and property [12].

The domain of explainable artificial intelligence (XAI) has recently gained traction. Various interpretability techniques have been proposed, and there are different methods to categorize XAI models, each focusing on specific parameters [13]. Some techniques operate at the individual level (IL), concentrating on the prediction for a single instance, while model-level (ML) or global-level models aim to describe the overall logic and behaviour of the models across the entire dataset. Certain models require training and learning another model, such as a surrogate model. Additionally, there are model-agnostic (MA) approaches that can be used with any type of machine learning model, irrespective of its internal structure, as well as model-specific (MS) techniques designed for specific model types. Some interpretability methods follow a backward flow, starting with the model's prediction and working backward to understand how the input features influenced these predictions, while others follow a forward flow [14].

While much progress has been made in XAI, particularly concerning DNNs for 2D images, there remains a lack of studies focused on the interpretability and explainability of 3D data, underscoring the need for further research in this area [15]. This study introduces a recent technique, SMILE [16], for use with point cloud data. SMILE, which has shown strong performance on 2D and tabular data, is derived from LIME [17] and involves generating perturbations of an instance from the dataset and training a simple, more interpretable surrogate model based on the distances and predictions of the main complex model. Although LIME has previously been applied to point cloud data [18], SMILE differs by utilizing Empirical Cumulative Distribution Function (ECDF)-based statistical distances, which offer greater stability compared to the cosine distances used in LIME.

The primary goal of this study is to enhance surrogate explainable models by introducing SMILE for point cloud data. The study's contributions include generating saliency maps for 3D objects, comparing the most important features influencing model predictions using LIME and various versions of SMILE under different parameter sets, and, also, evaluating the use of Bayesian Ridge as the surrogate model. The models are compared using fidelity parameters, and their stability is assessed by introducing noise into the point cloud object. Additionally, the study explores the classification of one of the most critical objects, the 'person' class, and examines its misclassified samples.

The structure of this study is as follows: Section 2 reviews related work, Section 3 outlines the algorithm and methodology of the proposed approach, Section 4 details the experimental results and discusses the findings, and Section 5 concludes the study.

## 2. State of the Art

As mentioned earlier, explainable models deployed in the realm of point clouds are relatively few. Table 1 provides a summary of these models, with this study's approach included, and compares them across different XAI parameters. The models are explained in more detail in this section.

PointHop was one of the pioneering attempts to make AI applied to point clouds more understandable [12]. Their work tackled the unordered nature of point clouds by employing PointHop units to make them compatible with traditional classifiers, serving as a pre-processing step rather than a post hoc explanation method. [19] proposed a method for generating saliency maps in point cloud recognition models by identifying non-contribution factors. This technique incorporates a different evaluation strategy that involves randomizing both the network weights and labels. In [18] the author employed a local surrogate model-based method to identify which factors play a role in classification. They also propose quantitative fidelity measurements for interpretability and compare the fidelity and plausibility of the surrogate model under different condition, as opposed to relying on a subjective approach relying on human judgment. In another study, [20] extended gradient-based methods including Vanilla Gradients, Guided Backpropagation, and Integrated Gradients to 3D point cloud and voxel data for the PointNet++ and Voxception-ResNet (VRN) models. They found that edges and corners are important features, while planar surfaces contribute less to classification decisions. Their results demonstrated that integrated gradients provided more uniform and meaningful attribution maps compared to the noisier vanilla gradients, especially for voxel-based inputs. In a novel framework named BubblEX, a new approach is proposed to enhance the explainability of deep learning models for 3D point-cloud classification [21]. The authors utilized t-SNE and UMAP techniques for dimensionality reduction to visualize high-dimensional feature data in a 2-D space and employed Grad-CAM for interpretability.

In another work, [22] introduces generative model-based Activation Maximization (AM) techniques to enhance the global explainability of point cloud neural networks, addressing the limitations of traditional methods in capturing the irregular and sparse structure of point clouds. Recently, [23] proposed DAM a novel method using Denoising Diffusion Probabilistic Models (DDPM) to generate high quality AM global explanations. Furthermore, DAM employs a Point Diffusion Transformer (PDT) with dual-classifier guidance, demonstrating superior performance in perceptibility, representativeness, and diversity, while also significantly reducing the time needed for explanation generation. In another recent approach, Feature Based Interpretability (FBI), a fast and simple XAI method for point cloud data are introduced [24]. Their method computes pointwise importance, allowing for better understanding of network properties. They demonstrated that pre-bottleneck computation is preferred over post-bottleneck, offering smoother and higher quality importance ranking. The FBI method is significantly faster than other XAI methods, making it scalable for large point clouds and extensive network architectures [24].

*Table 1: Summary of XAI models for point cloud data*

| Method | Ex. Type | Learning | Task | Approach | Flow | Dataset |
| --- | --- | --- | --- | --- | --- | --- |
| PointHop [12] | ML | Yes | PC | MS | Forward | Modelnet40 [7] |
| Non-Contribution Factors [19] | IL | No | PC | MS | Backward | Modelnet40 [7] |
| LIME [18] | IL | Yes | PC | MA | Forward | Modelnet40 [7] |
| Gradient-Based [20] | IL | No | PC/VD | MS | Backward | Modelnet40 [7] |
| BubblEX [21] | IL | No | PC | MA | Backward | Modelnet40 [7] ScanObjectNN [10] |
| AM [22] | ML | Yes | PC | MS | Backward | ModelNet40 [7] ShapeNet [8] |
| DAM [23] | ML | Yes | PC | MS | Backward | ModelNet40 [7] ShapeNet [8] |
| FBI [24] | IL | No | PC | MA | Forward | ModelNet40 [7] ModelNet-C [11] ScanObjectNN [10] |
| **SMILE (Our Method)** | IL | Yes | PC | MA | Forward | ModelNet40 [7] |

# 3. Methodology

## 3.1. Implementation Details

The primary difference between LIME and SMILE lies in their distance calculation methods: SMILE uses Empirical Cumulative Distribution Function-based (ECDF) statistical distances, whereas LIME employs Cosine distances [16]. As depicted in Figure 1, these stages include generating randomly perturbed inputs around a given sample, predicting labels using a black-box classifier, calculating distances between the sample and each perturbation, determining feature weights using a kernel function, and deploying a weighted linear regression as a surrogate model.

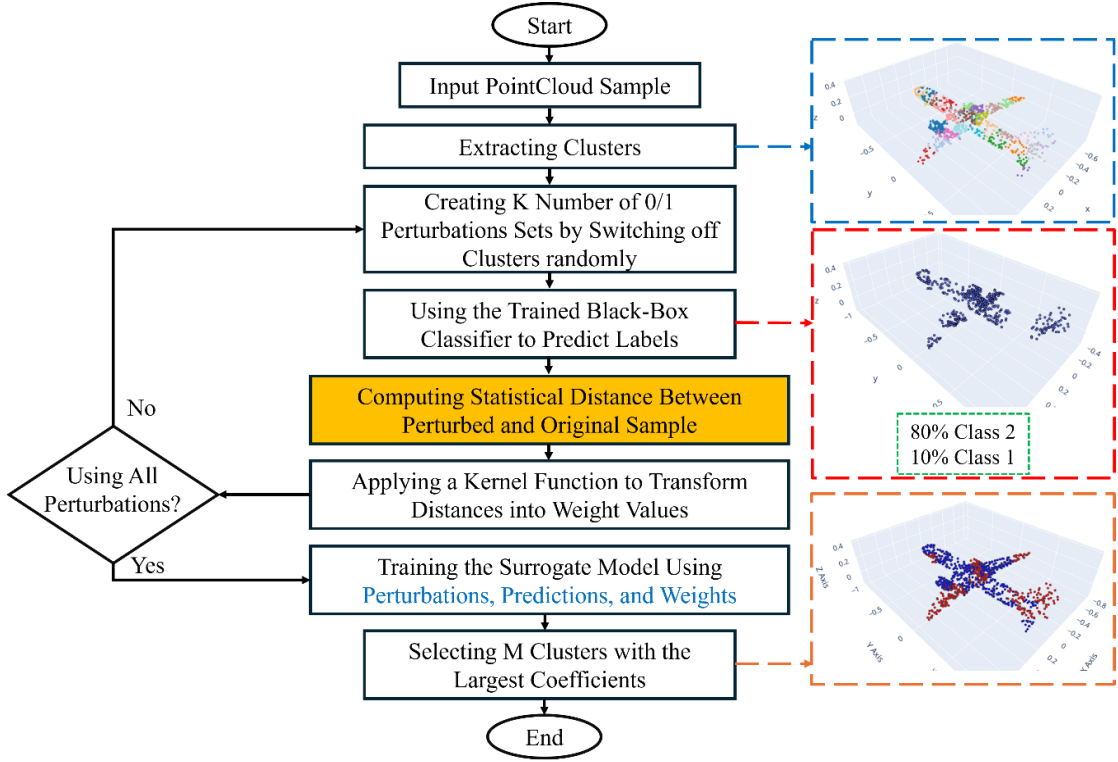

Figure 1: SMILE flowchart for explaining point cloud classification

To manage computational complexity when explaining point cloud data, the points are clustered into super points using 3D K-Means clustering and Farthest Point Sampling (FPS) [18]. The core concept of FPS involves iteratively choosing the point that is most distant from the previously selected points. These chosen points create a subset that closely represents the original set, reducing the total number of points while retaining the key features of the original data [25].

Perturbations are then generated using a binary mask matrix, The mask perturbation matrix $M$ is generated using a binomial distribution with a success probability of 0.5, ensuring each cluster has an equal chance of being turned on or off independently. These perturbations are fed into the black-box model to generate class probabilities.

In SMILE, ECDF statistical distances, such as Wasserstein Distance (WD), are calculated to capture the geometry-related features of the distributions between the main instance and perturbations. The WD calculates the distance between two univariate distributions $F_1(x)$ and $F_2(x)$ as follows [26]:

$$WD = \int_{-\infty}^{+\infty} |F_1(x) - F_2(x)| \, dx \qquad (1)$$

Here, $F_1(x)$ represents the ECDF of the target point and its surrounding local distributed samples, while $F_2(x)$ represents the ECDF of the clustered random points relative to their class labels. Other ECDF-based measures, including Kolmogorov-Smirnov and Anderson-Darling distances, are also considered [16].

The total distance across all coordinates is computed, and the resulting distances are mapped to weights using an exponential smoothing kernel function:

$$w_i = \exp\left(-\frac{WD^2}{\sigma^2}\right) \tag{2}$$

The kernel width $\sigma$ is crucial in determining the sensitivity of the weights, with a range of $\sigma$ values explored for optimal results.

Finally, a simple model is trained using the features, predicted labels, and calculated weights. The model's coefficients provide insights into the impact of each feature on the different classes.

### 3.2. Evaluation Metrics

In evaluating the performance and reliability of the explainable model in relation to the black-box model, fidelity and stability metrics are employed. These metrics offer deeper insight into the models' performance beyond traditional methods like saliency maps and human inspection.

#### i. Fidelity:

To measure the alignment between the black-box model and the explainable model, fidelity is employed as a metric. Given $f(Z_i)$ and $g(Z_i)$ as their respective predictions for the $i$-th perturbation, various loss and coefficient metrics assess the similarity between the regression scores and predictions. One such metric is mean loss, defined as [18]:

$$L_m = \left| \sum_{i=1}^{N_p} \left(\frac{f(Z_i)}{N_p}\right) - \sum_{i=1}^{N_p} \left(\frac{g(Z_i)}{N_p}\right) \right| \tag{3}$$

Additional metrics include the Mean $L_1$ and $L_2$ losses:

$$L_1 = \frac{1}{N_p} \sum_{i=1}^{N_p} |f(Z_i) - g(Z_i)| \tag{4}$$

$$L_2 = \frac{1}{N_p} \sum_{i=1}^{N_p} (f(Z_i) - g(Z_i))^2 \tag{5}$$

Weighted versions of these losses are also used:

$$L_1^w = \frac{1}{N_p} \sum_{i=1}^{N_p} (|f(Z_i) - g(Z_i)| \cdot w) \tag{6}$$

$$L_2^w = \frac{1}{N_p} \sum_{i=1}^{N_p} ((f(Z_i) - g(Z_i))^2 \cdot w) \tag{7}$$

The weighted coefficient of determination is given by:

$$R_w^2 = 1 - \frac{\sum_{i=1}^{N_p}(f(Z_i) - g(Z_i))^2}{\sum_{i=1}^{N_p}(f(Z_i) - \overline{f_w(Z_i)})^2} \tag{8}$$

Moreover, the weighted adjusted coefficient of determination is defined as:

$$\hat{R}_w^2 = 1 - (1 - R_w^2)\left[\frac{N_p - 1}{N_p - N_s - 1}\right] \qquad (9)$$

where $w$ represents kernel weights, and $\overline{f_w(Z_i)}$ is the weighted average.

Discrepancies in predicted scores are quantified using $L_m$, $L_1^w$ and $L_2^w$, while $R_w^2$ evaluates the correlation between the proxy model and the neural network. Given that $R_w^2$ is susceptible to positive bias in small sample sizes, $\hat{R}_w^2$ adjusts for sample size and variable count [27].

ii. **Stability:**

An explainable model should exhibit stability, meaning that minor changes to the input should not affect the model's predictions or explanations [28]. An explanation $m_i$ for an input graph $g_i$ is regarded as the ground truth. The input graph $g_i$ is then perturbed by minor changes, such as adding new cluster of points, resulting a new graph $g'_i$ having the same prediction. The explanations for $g'_i$ are obtained and denoted as $m'_i$. Therefore, by comparing the differences between $m_i$ and $m'_i$ the stability can be calculated.

Previous research has highlighted that local surrogate models may be unstable across various data types, including text, tabular, and image data [29]. This study extends stability analysis to point cloud data, employing the Jaccard index as the metric. The Jaccard index, defined as:

$$Jaccard(A, B) = \frac{|A \cap B|}{|A \cup B|} \qquad (10)$$

measures the similarity between two sets $A$ and $B$ as the ratio of their intersection to their union with values bounded between 0 and 1.

## 4. Experiments and Discussion

While the proposed SMILE model is model-agnostic and can be extended to any point cloud model, this study examines one of the most well-known models, PointNet [30]. PointNet, based on pointwise MLP networks, achieved an accuracy of 89.2% on the ModelNet40 dataset. A previous study [18] used this model to demonstrate the capability of LIME on 3D point cloud data, making it as the only work in the literature that has applied an XAI model to the ModelNet40 dataset in a manner directly comparable to our study. Additionally, Kernel SHAP (SHapley Additive exPlanations) [31] was explored for a more comprehensive comparison between XAI methods. Although specific objects and results from the referenced study were not accessible, the LIME and Kernel SHAP results are referenced here due to alignment with their cited methodologies.

### 4.1. Cluster Number Comparison

To obtain the saliency map and fidelity measurements, an object from the airplane class is selected from the dataset and converted to point cloud data with 1024 points. The object is then examined under different cluster numbers (32, 64, 128, and 1024) using the K-means clustering algorithm along with FPS. The number of perturbations is set to 1000, with weighted regression used as the surrogate model. After sorting the clusters by importance, the top 20% of the most significant ones are highlighted in red to indicate the critical features used by the main model for classification.

As depicted in Figure 2, the saliency maps for LIME and SMILE are identical. Both methods highlight the tail and wings of the airplane as the most important features for predicting the labels, which aligns logically with human intuition. However, it is noteworthy that when the number of clusters is set to 1024, both models fail to identify the most important features. This indicates that for this number of clusters, the number of perturbations should be increased to achieve suitable results [18]. In contrast, SHAP's saliency maps become inconsistent as cluster numbers increase, suggesting it may require more perturbations or adjustments for point cloud data.

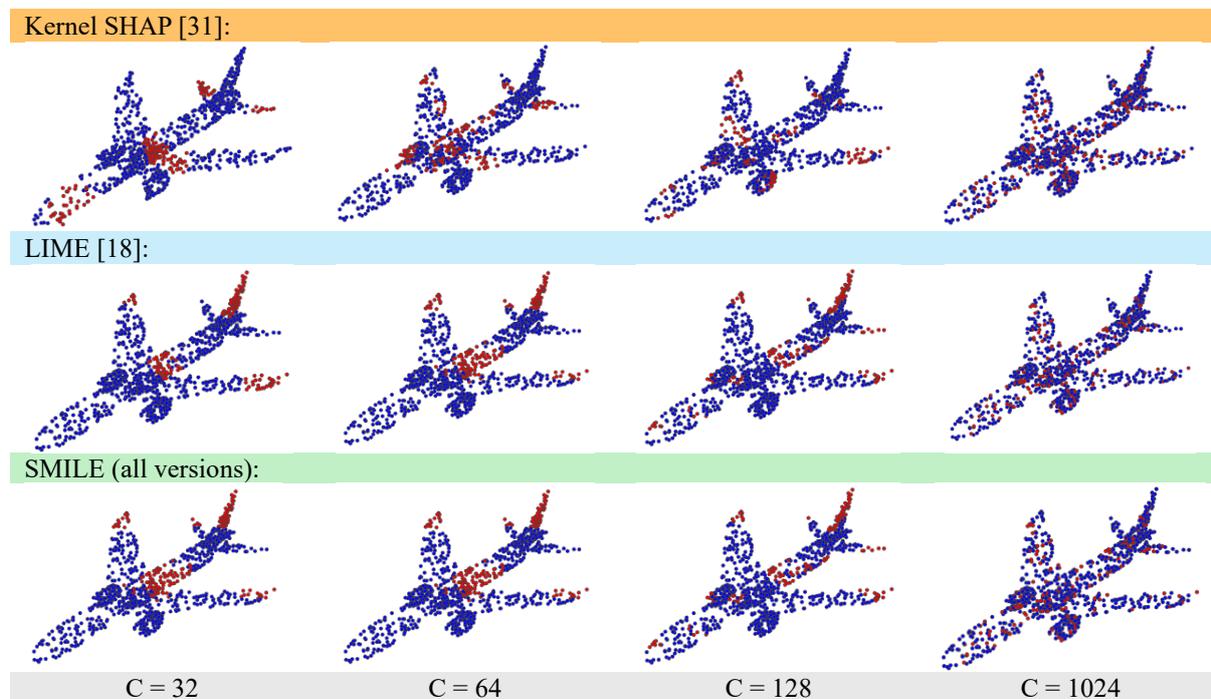

Figure 2: Saliency Maps obtained by LIME [18] and SMILE-WD with different numbers of clusters

To enable a more comprehensive comparison, fidelity parameters are employed to evaluate the SMILE model using ECDF statistical distances, including Wasserstein (SMILE-WD), Anderson-Darling (SMILE-AD), and Kolmogorov-Smirnov (SMILE-KS) distances, alongside LIME, which uses cosine distance. The results, summarized in Table 2, indicate that all SMILE methods generally exhibit higher weighted ($R_w^2$) and weighted adjusted ($\hat{R}_w^2$) coefficients, suggesting superior performance. Furthermore, SMILE-AD demonstrates significantly lower mean loss compared to all other methods and also attains the highest $\hat{R}_w^2$ performance among the methods. However, as the number of clusters increases, both $R_w^2$, $\hat{R}_w^2$ decrease across all models, suggesting that the accuracy of the explainable models diminishes with higher cluster counts. This trend indicates a potential need to increase the number of perturbations when working with higher cluster sizes to maintain accuracy.

Regarding $L_1, L_2, L_1^w, L_2^w$, some inconsistencies in the results are observed, which stem from the nature of the comparisons. Specifically, the formulation compares the probabilities generated by the models rather than the predicted classes. As these parameters compare probabilities on a one-to-one basis, it can lead to inconsistencies in the results.

*Table 2: Local fidelity of various explanation methods and cluster sizes*

|  | C | $L_m$ | $L_1$ | $L_1^w$ | $L_2$ | $L_2^w$ | $R_w^2$ | $\hat{R}_w^2$ |
|---|---|---|---|---|---|---|---|---|
| LIME [18] | 32 | $7.69 \times 10^{-4}$ | 0.132 | $1.09 \times 10^{-1}$ | $2.30 \times 10^{-2}$ | $2.42 \times 10^{-2}$ | 0.529 | 0.513 |
| SMILE-WD | 32 | $1.24 \times 10^{-3}$ | 0.132 | $1.26 \times 10^{-1}$ | $2.30 \times 10^{-2}$ | $2.80 \times 10^{-2}$ | 0.536 | 0.521 |
| SMILE-AD | 32 | $\mathbf{6.83 \times 10^{-9}}$ | 0.133 | $1.05 \times 10^{-9}$ | $2.30 \times 10^{-2}$ | $2.97 \times 10^{-10}$ | **0.539** | **0.524** |
| SMILE-KS | 32 | $3.83 \times 10^{-4}$ | 0.133 | $1.03 \times 10^{-1}$ | $2.30 \times 10^{-2}$ | $2.92 \times 10^{-2}$ | 0.538 | 0.523 |
| LIME [18] | 64 | $1.37 \times 10^{-4}$ | 0.047 | $1.84 \times 10^{-2}$ | $6.65 \times 10^{-3}$ | $5.23 \times 10^{-3}$ | 0.300 | 0.252 |
| SMILE-WD | 64 | $2.81 \times 10^{-4}$ | 0.047 | $2.22 \times 10^{-2}$ | $6.65 \times 10^{-3}$ | $6.37 \times 10^{-3}$ | 0.305 | 0.257 |
| SMILE-AD | 64 | $\mathbf{7.66 \times 10^{-8}}$ | 0.047 | $2.3 \times 10^{-10}$ | $6.65 \times 10^{-3}$ | $6.65 \times 10^{-11}$ | **0.307** | **0.260** |
| SMILE-KS | 64 | $7.00 \times 10^{-5}$ | 0.047 | $2.30 \times 10^{-2}$ | $6.65 \times 10^{-3}$ | $6.57 \times 10^{-3}$ | 0.306 | **0.260** |
| LIME [18] | 128 | $1.86 \times 10^{-6}$ | 0.010 | $2.59 \times 10^{-3}$ | $4.66 \times 10^{-4}$ | $3.87 \times 10^{-4}$ | 0.285 | 0.180 |
| SMILE-WD | 128 | $2.14 \times 10^{-5}$ | 0.010 | $3.11 \times 10^{-3}$ | $4.66 \times 10^{-4}$ | $4.55 \times 10^{-4}$ | 0.286 | 0.181 |
| SMILE-AD | 128 | $\mathbf{7.56 \times 10^{-9}}$ | 0.010 | $3.2 \times 10^{-11}$ | $4.66 \times 10^{-4}$ | $4.66 \times 10^{-12}$ | **0.288** | **0.183** |
| SMILE-KS | 128 | $4.58 \times 10^{-7}$ | 0.010 | $3.17 \times 10^{-3}$ | $4.66 \times 10^{-4}$ | $4.64 \times 10^{-4}$ | **0.288** | **0.183** |

### 4.2. Kernel width Comparison

To evaluate the impact of kernel width on model performance, a range from 0.1 to 0.7, with 0.1 increments, was tested. Figure 3 shows the saliency maps generated by SMILE and LIME across different kernel widths. Notably, all SMILE variants produced identical saliency maps, indicating consistency across the methods.

Regarding fidelity, as shown in Figure 4 increasing kernel width generally reduced Mean Loss and improved the $\hat{R}_w^2$ score, reflecting enhanced performance. SMILE-AD, using the Anderson-Darling distance, consistently maintained a high $\hat{R}_w^2$ score of 0.54 and a Mean Loss in the order of $10^{-9}$, demonstrating superior stability compared to LIME. This stability, likely due to the AD distance's sensitivity to the tails of distributions, indicates that SMILE-AD provides consistent explanations without the need for fine-tuning the kernel width.

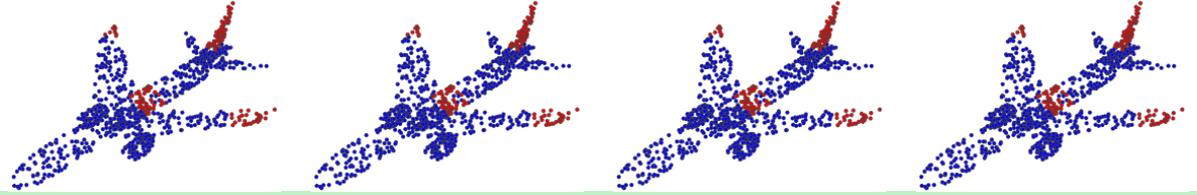

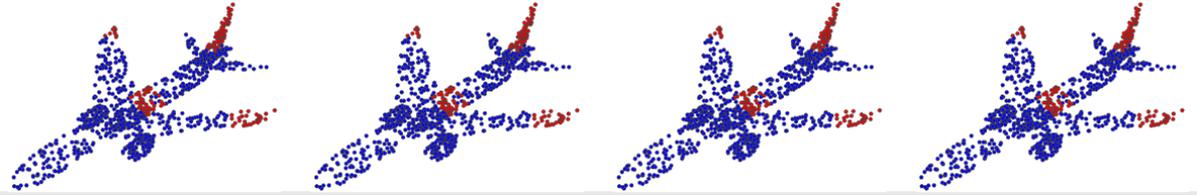

|  |  |  |  |
|---|---|---|---|
| $\sigma = 0.1$ | $\sigma = 0.2$ | $\sigma = 0.3$ | $\sigma = 0.70$ |

Figure 3: Saliency Maps obtained by LIME [18] and SMILE with Different kernel widths

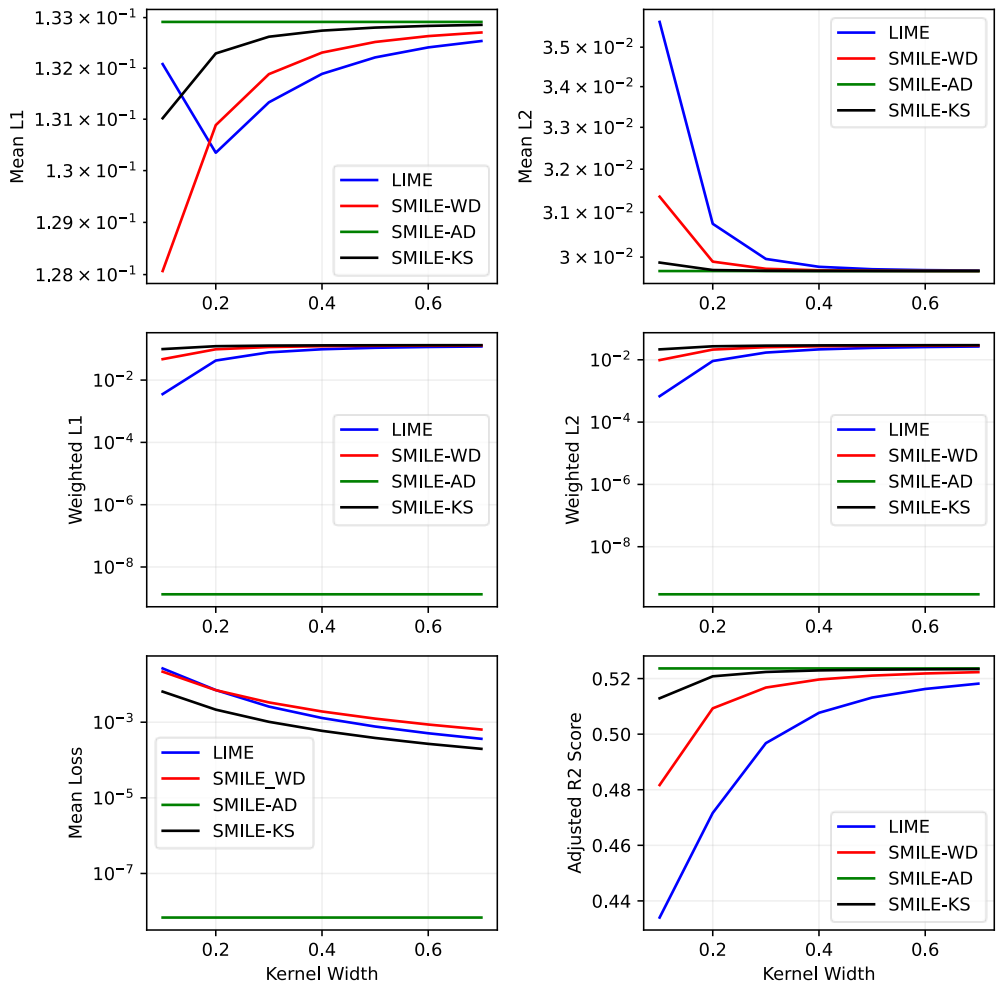

Figure 4: Performance comparison of LIME [18] and SMILE across kernel width (32 clusters) – losses on a logarithmic scale

### 4.3. Perturbation Number Comparison

Another examination was conducted to assess the effect of the number of perturbations on approach model, using a cluster count of 32 and a kernel width of 0.5. The fidelity scores of models are compared in Figure 5. The loss metrics fluctuated without significant differences across models when compared to themselves. However, $\hat{R}_w^2$ generally decreased with higher perturbations, except for a surprising increase at 450 perturbations. This trend occurs because fidelity scores compare probabilities, making them sensitive to the number of perturbations. In other words, as the number of perturbations increases, more comparisons are made, leading to higher loss and, consequently, lower $\hat{R}_w^2$.

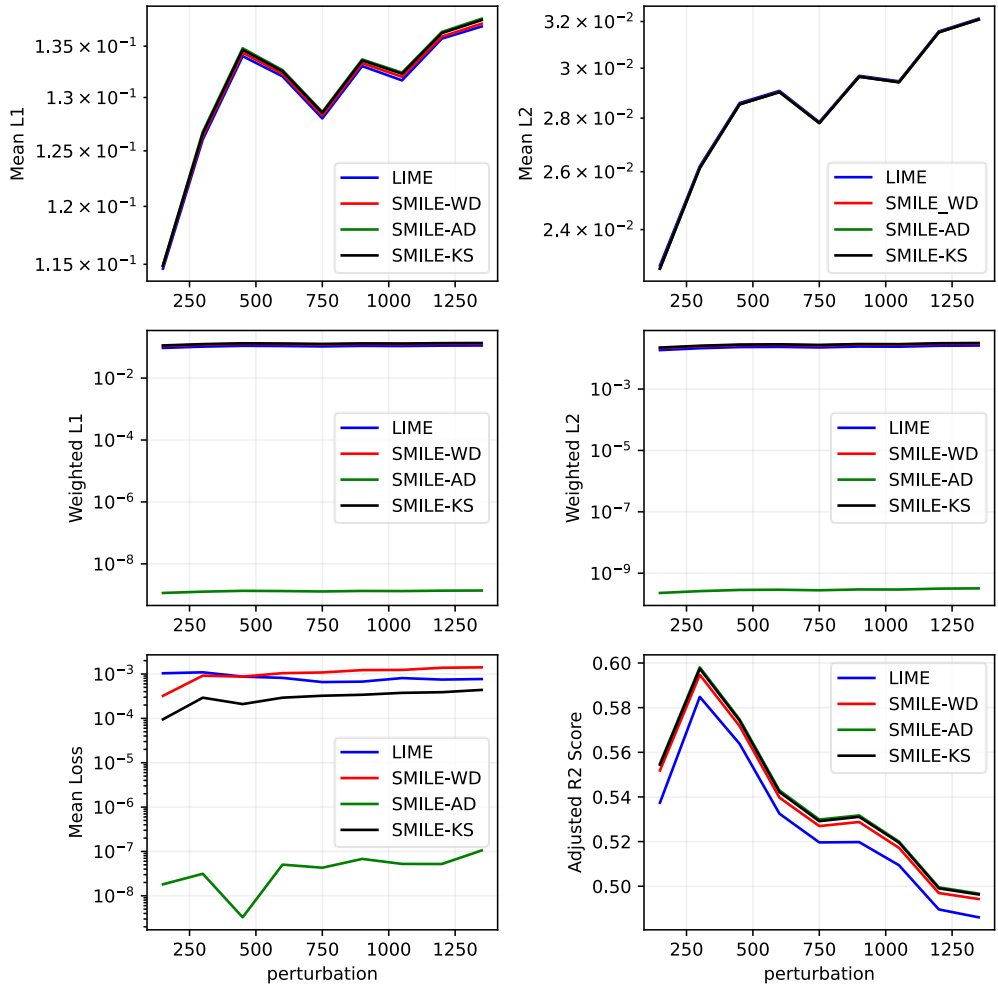

Figure 5: Performance comparison of LIME [18] and SMILE across the number of perturbations (32 clusters) – losses on a logarithmic scale

When comparing saliency maps across different perturbation counts, the maps for all models were identical. Therefore, only a single SMILE map is presented in Figure 6. The saliency maps varied among models with fewer than 750 perturbations, but remained consistent beyond this threshold, indicating that a minimum of 750 perturbations is sufficient for stable results with 32 clusters. This suggests an acceptable baseline for perturbations in this setting.

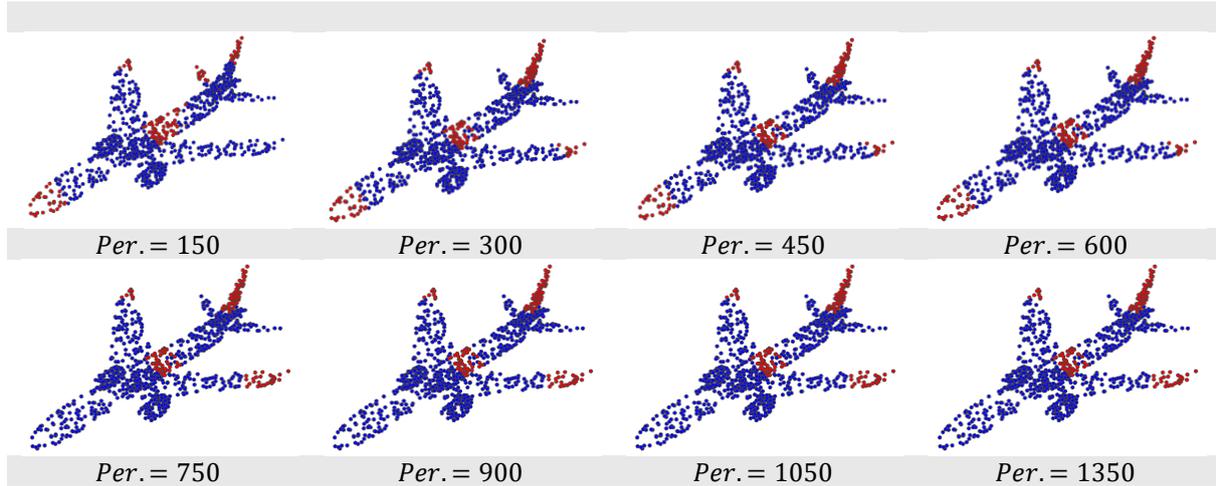

Figure 6: Saliency maps obtained by SMILE with varying perturbation numbers

### 4.4. Surrogate Model Comparison

In another experiment, LIME, SMILE-WD, SMILE-AD, and SMILE-KS were evaluated using Bayesian Ridge as the surrogate model instead of weighted regression, with a kernel width of 0.5, 32 clusters, and 1000 perturbations. Comparing the results in Table 3 with those in Table 2 indicates that $\hat{R}_w^2$ values for all SMILE variants and LIME were slightly higher with weighted regression compared to Bayesian Ridge. Notably, SMILE-AD, which was the best model with weighted regression, experienced a significant drop in $\hat{R}_w^2$ with Bayesian Ridge, making it the worst-performing model among the others.

*Table 3: Local fidelity of various explanation methods using Bayesian Ridge as the surrogate model*

|  | C | $L_m$ | $L_1$ | $L_1^w$ | $L_2$ | $L_2^w$ | $R_w^2$ | $\hat{R}_w^2$ |
|---|---|---|---|---|---|---|---|---|
| LIME [18] | 32 | $1.00 \times 10^{-3}$ | 0.132 | 0.129 | 0.030 | 0.029 | 0.528 | 0.512 |
| SMILE-WD | 32 | $1.29 \times 10^{-3}$ | 0.132 | 0.131 | 0.030 | 0.029 | 0.536 | 0.520 |
| SMILE-AD | 32 | $6.83 \times 10^{-9}$ | 0.131 | 0.131 | 0.032 | 0.032 | 0.513 | 0.496 |
| SMILE-KS | 32 | $3.94 \times 10^{-4}$ | 0.132 | 0.132 | 0.030 | 0.030 | 0.538 | 0.523 |

### 4.5. Stability

To further evaluate the stability of the proposed model, a cluster of 30 points, shaped like a ball, was randomly inserted into various locations on the object. LIME and SMILE were then employed to identify the clusters with the most significant contribution to the model's predictions. As shown in Figure 7, critical features, including clusters on the tail and wings of the airplane, generally remained stable even after the introduction of the bulk. However, the importance of features shifted noticeably when the bulk itself was identified as the most significant feature. This shift likely occurs because the model prioritizes the bulk when it is positioned near crucial features, indicating the model's reliance on the spatial proximity of points within the point cloud.

By comparing 10 perturbed samples using the Jaccard index, a mean value of 0.78 was obtained for both the LIME and SMILE methods. This value suggests a moderate level of stability, indicating that while the model's explanations are somewhat influenced by perturbations, the overall consistency remains acceptable. These results highlight the robustness of the model and the reliability of explanations provided by both LIME and SMILE, even with minor input changes.

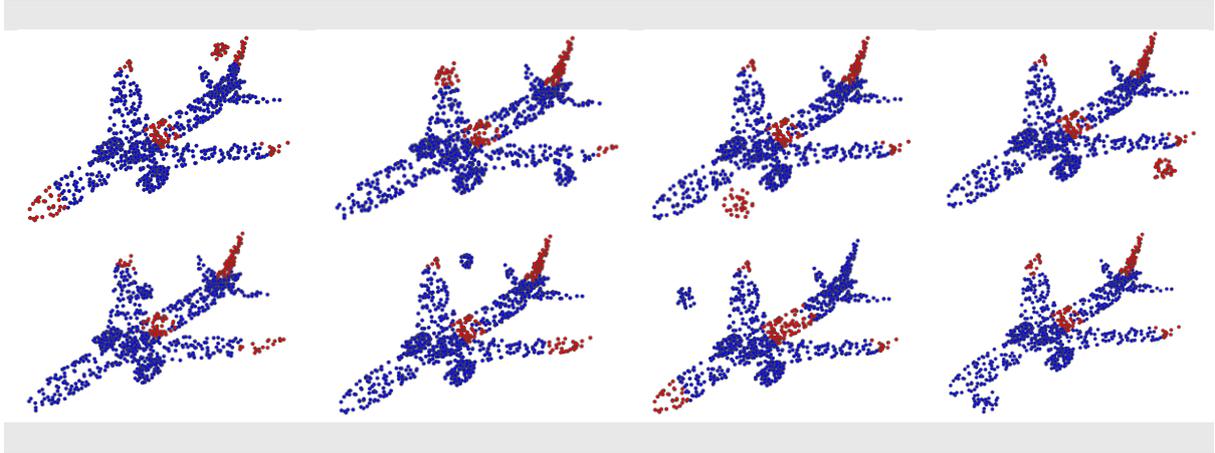

Figure 7: Saliency maps of a point cloud object with noise inserted in different regions of the data.

### 4.6. Computation Complexity

Since the number of perturbations directly influences running time, a fixed 1000 perturbations were used to compare the running times of all models with 32, 64, and 128 clusters, using either weighted regression or Bayesian Ridge as the surrogate model. As shown in Figure 8, models using Bayesian Ridge exhibit slightly shorter running times compared to those using weighted regression. Among the models, LIME is the most time-efficient, while SMILE-KS is the least, with a difference of approximately 3 seconds.

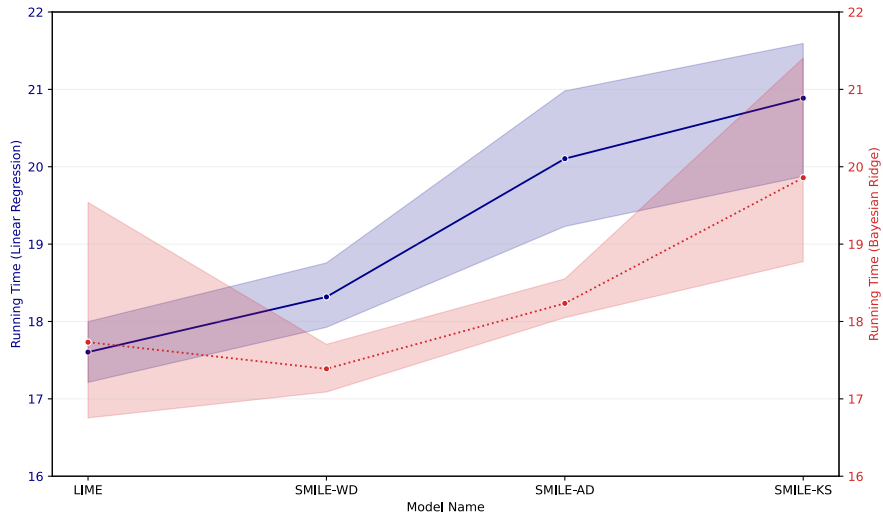

Figure 8: Comparison of running time between models

Table 4 presents a comparison of the running time of our approach against other methods that provide global explanations for PointNet and PointNet++ [22, 23]. While a direct comparison is challenging due to the unspecified hardware configurations and differing test objects in these studies, the table serves to illustrate the general order of running times. As shown, all surrogate models, which were executed on an Intel(R) Core (TM) i5-10505 CPU @ 3.20GHz, demonstrate efficient time consumption compared to other approaches. It is important to note that number of perturbations in our experiments was set to 1000, although perturbation analysis indicates that satisfactory results can be achieved with fewer perturbations. Reducing the number of perturbations would further decrease the running time, making it potentially comparable to DAM, which is recognized as one of the fastest explainable models for point cloud data [23].

*Table 4: Comparative analysis of model running time*

|         | AE [22] | AED [22] | NAED [22] | DAM [23] | LIME [18] | SMILE-WD | SMILE-AD | SMILE-KS |
|---------|---------|----------|-----------|----------|-----------|----------|----------|----------|
| Time(s) | 47.75   | 458.69   | 201.27    | 12.35    | 16.75     | 17.08    | 18.55    | 18.77    |

### 4.7. SMILE Analysis of Misclassified Objects

Failure analysis is a crucial application of explainable models, as it aids in identifying incorrect areas of focus by the classifier, thereby providing insights for potential model or dataset improvements. In this study, the 'person' object was selected as one of the most important categories for classification, given its significance in improving the reliability and safety of autonomous robots and vehicles. The ModelNet40 dataset [7] contains only 18 objects of this class in its test set, with PointNet correctly predicting 11 of them [30]. However, a more detailed examination of the dataset revealed that most 'person' objects are depicted holding items resembling a sword or gun. Notably, the model struggled to correctly classify those 'person' objects that displayed a normal gesture without any additional objects. Using SMILE to identify the most important features the model focuses on for correctly labelled 'person' objects, as shown in Figure 9, indicates that the presence of objects like a gun or sword, along with the person, are considered critical features by the model. This suggests that the model incorrectly prioritizes these additional objects, leading to misclassification when they are absent.

This analysis suggests that ModelNet40 may not be an ideal dataset for detecting humans. Future models intended for point cloud classification should consider exploring other available datasets or augmenting ModelNet40 with a more comprehensive set of 'person' objects.

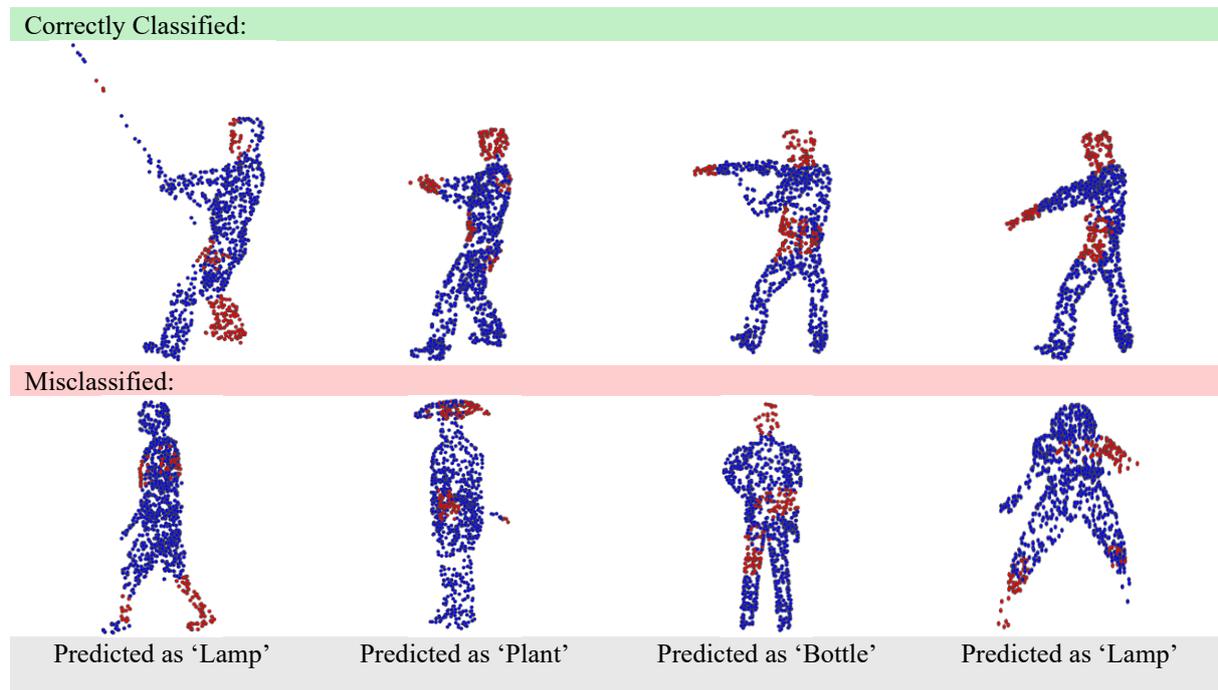

Figure 9: Saliency maps for correctly classified and misclassified 'person' class objects by PointNet, generated using SMILE.

## 5. Conclusion

In response to the increasing demand for more interpretable and explainable AI models, this study evaluates the novel SMILE approach, a refinement of LIME that uses Empirical Cumulative Distribution Function (ECDF) statistical distances, within the context of point cloud data. Comparative analysis with LIME reveals that SMILE consistently offers superior fidelity scores, particularly with the Anderson-Darling distance (SMILE-AD), which demonstrates robust stability unaffected by kernel width. Perturbation analysis indicates that the surrogate models remain stable with over 700 perturbations and 32 clusters, and linear regression outperforms Bayesian Ridge as a surrogate model in terms of fidelity. Although LIME is more computationally efficient, all tested models operate within acceptable time limits. Stability assessments, indicated by a Jaccard index of 0.78, affirm a suitable level of stability for the SMILE approach, setting a new benchmark for measuring the stability of point cloud models and establishing a standard for future research in this area.

Moreover, the application of SMILE to classify 'person' objects exposed dataset biases, highlighting the need for more comprehensive and contextually relevant data, particularly for critical applications such as autonomous driving and robotics.

This study underscores the importance of further research into both explainability methods and dataset improvements to enhance model interpretability and robustness. There is significant potential for advancing explainability in point cloud DNNs through the exploration of alternative surrogate models and other explainable approaches.

## Code Availability

Regarding the research reproducibility, codes and functions supporting this paper are published online at GitHub: https://github.com/Dependable-Intelligent-Systems-Lab/xwhy.

## Acknowledgments

The financial support from the Secure and Safe Multi-Robot Systems (SESAME) H2020 Project, under Grant Agreement 101017258, is gratefully acknowledged. Further appreciation is extended to the University of Hull and Dock Robotics Ltd for their invaluable technical and administrative contributions.

# Bibliography

1. Heer, C. *World Robotics Report: "All-Time High" with Half a Million Robots Installed in one Year*. 2022 [cited 2024; Available from: https://ifr.org/ifr-press-releases/news/wr-report-all-time-high-with-half-a-million-robots-installed.
2. Guo, Y., et al., *Deep learning for 3d point clouds: A survey.* IEEE transactions on pattern analysis and machine intelligence, 2020. **43**(12): p. 4338-4364.
3. Xiao, A., et al., *Unsupervised Point Cloud Representation Learning With Deep Neural Networks: A Survey.* IEEE Transactions on Pattern Analysis and Machine Intelligence, 2023. **45**(9): p. 11321-11339.
4. Muzahid, A.A.M., et al., *Deep learning for 3D object recognition: A survey.* Neurocomputing, 2024. **608**: p. 128436.
5. Siddiqi, K., et al., *Retrieving articulated 3-D models using medial surfaces.* Mach. Vis. Appl., 2008. **19**: p. 261-275.
6. Deuge, M., et al., *Unsupervised feature learning for classification of outdoor 3D Scans.* Australasian Conference on Robotics and Automation, ACRA, 2013.
7. Zhirong, W., et al. *3D ShapeNets: A deep representation for volumetric shapes*. in *2015 IEEE Conference on Computer Vision and Pattern Recognition (CVPR)*. 2015.
8. Chang, A.X., et al., *Shapenet: An information-rich 3d model repository.* arXiv preprint arXiv:1512.03012, 2015.
9. Dai, A., et al. *Scannet: Richly-annotated 3d reconstructions of indoor scenes*. in *Proceedings of the IEEE conference on computer vision and pattern recognition*. 2017.
10. Uy, M.A., et al. *Revisiting point cloud classification: A new benchmark dataset and classification model on real-world data*. in *Proceedings of the IEEE/CVF international conference on computer vision*. 2019.
11. Ren, J., L. Pan, and Z. Liu. *Benchmarking and analyzing point cloud classification under corruptions*. in *International Conference on Machine Learning*. 2022. PMLR.
12. Zhang, M., et al., *Pointhop: An explainable machine learning method for point cloud classification.* IEEE Transactions on Multimedia, 2020. **22**(7): p. 1744-1755.
13. Mersha, M., et al., *Explainable artificial intelligence: A survey of needs, techniques, applications, and future direction.* Neurocomputing, 2024. **599**: p. 128111.
14. Alicioglu, G. and B. Sun, *A survey of visual analytics for Explainable Artificial Intelligence methods.* Computers & Graphics, 2022. **102**: p. 502-520.
15. Zhao, B., et al., *Evaluation of Convolution Operation Based on the Interpretation of Deep Learning on 3D Point Cloud.* IEEE Journal of Selected Topics in Applied Earth Observations and Remote Sensing, 2020. **PP**: p. 1-1.
16. Aslansefat, K., et al., *Explaining black boxes with a SMILE: Statistical Model-agnostic Interpretability with Local Explanations.* IEEE Software, 2023.
17. Ribeiro, M.T., S. Singh, and C. Guestrin. *" Why should i trust you?" Explaining the predictions of any classifier*. in *Proceedings of the 22nd ACM SIGKDD international conference on knowledge discovery and data mining*. 2016.
18. Tan, H. and H. Kotthaus. *Surrogate model-based explainability methods for point cloud nns*. in *Proceedings of the IEEE/CVF Winter Conference on Applications of Computer Vision*. 2022.
19. Liang, A., H. Zhang, and H. Hua, *Point Cloud Saliency Maps Based on Non-Contribution Factors*. 2022. 194-198.
20. Gupta, A., S. Watson, and H. Yin. *3d point cloud feature explanations using gradient-based methods*. in *2020 International Joint Conference on Neural Networks (IJCNN)*. 2020. IEEE.
21. Matrone, F., et al., *BubblEX: An Explainable Deep Learning Framework for Point-Cloud Classification.* IEEE Journal of Selected Topics in Applied Earth Observations and Remote Sensing, 2022. **15**: p. 1-18.
22. Tan, H. *Visualizing global explanations of point cloud dnns*. in *Proceedings of the IEEE/CVF Winter Conference on Applications of Computer Vision*. 2023.
23. Tan, H., *DAM: Diffusion Activation Maximization for 3D Global Explanations.* arXiv preprint arXiv:2401.14938, 2024.

24. Levi, M.Y. and G. Gilboa, *Fast and Simple Explainability for Point Cloud Networks.* arXiv preprint arXiv:2403.07706, 2024.
25. Qi, C.R., et al., *Pointnet++: Deep hierarchical feature learning on point sets in a metric space.* Advances in neural information processing systems, 2017. **30**.
26. Aslansefat, K., et al., *Toward Improving Confidence in Autonomous Vehicle Software: A Study on Traffic Sign Recognition Systems.* Computer, 2021. **54**(8): p. 66-76.
27. Ali, M.A., *Effect of Sample Size on the Size of the Coefficient of Determination in Simple Linear Regression.* Journal of Information and Optimization Sciences, 1987. **8**: p. 209-219.
28. Yuan, H., et al., *Explainability in graph neural networks: A taxonomic survey.* IEEE transactions on pattern analysis and machine intelligence, 2022. **45**(5): p. 5782-5799.
29. Burger, C., C. Walter, and T. Le, *The Effect of Similarity Measures on Accurate Stability Estimates for Local Surrogate Models in Text-based Explainable AI.* arXiv preprint arXiv:2406.15839, 2024.
30. Qi, C.R., et al. *Pointnet: Deep learning on point sets for 3d classification and segmentation.* in *Proceedings of the IEEE conference on computer vision and pattern recognition.* 2017.
31. Aas, K., M. Jullum, and A. Løland, *Explaining individual predictions when features are dependent: More accurate approximations to Shapley values.* Artificial Intelligence, 2021. **298**: p. 103502.